# Automatic Image Segmentation by Dynamic Region Merging

Bo Peng, Lei Zhang[1], *Member, IEEE* and David Zhang, *Fellow Member, IEEE*

Department of Computing, The Hong Kong Polytechnic University, Hong Kong

**Abstract:** This paper addresses the automatic image segmentation problem in a region merging style. With an initially over-segmented image, in which the many regions (or super-pixels) with homogeneous color are detected, image segmentation is performed by iteratively merging the regions according to a statistical test. There are two essential issues in a region merging algorithm: order of merging and the stopping criterion. In the proposed algorithm, these two issues are solved by a novel predicate, which is defined by the sequential probability ratio test (SPRT) and the maximum likelihood criterion. Starting from an over-segmented image, neighboring regions are progressively merged if there is an evidence for merging according to this predicate. We show that the merging order follows the principle of dynamic programming. This formulates image segmentation as an inference problem, where the final segmentation is established based on the observed image. We also prove that the produced segmentation satisfies certain global properties. In addition, a faster algorithm is developed to accelerate the region merging process, which maintains a *nearest neighbor graph* in each iteration. Experiments on real natural images are conducted to demonstrate the performance of the proposed dynamic region merging algorithm.

---

[1] Corresponding author: cslzhang@comp.polyu.edu.hk. This work is supported by the Hong Kong SAR General Research Fund (PolyU 5330/07E).

# I. Introduction

Image segmentation is a fundamental yet still challenging problem in computer vision and image processing. In particular, it is an essential process for many applications such as object recognition, target tracking, content-based image retrieval and medical image processing, etc. Generally speaking, the goal of image segmentation is to partition an image into a certain number of pieces which have coherent features (color, texture, etc.) and in the meanwhile to group the meaningful pieces together for the convenience of perceiving [1]. In many practical applications, as a large number of images are needed to be handled, human interactions involved in the segmentation process should be as less as possible. This makes automatic image segmentation techniques more appealing. Moreover, the success of many high-level segmentation techniques (e.g. class-based object segmentation [2, 3]) also demands sophisticated automatic segmentation techniques.

Dating back over decades, there is a large amount of literature on automatic image segmentation. For example, some classical algorithms are based on the abrupt changes in image intensity or color, thus salient edges can be detected. Well-known edge detection algorithms include the Robert edge detector, the Sobel edge detector, the Laplacian operator [4] and the Canny edge detector [5-7]. However, due to the resulting edges are often discontinuous or over-detected, they can only provide candidates for the object boundaries. As the image segmentation problems can be easily translated into graph-related problems, some algorithms were proposed based on the criterion of minimizing the cut on a graph. The most widely used cut criteria include normalized cut [8], ratio cut [9] and minimum cut [10], in which the cost functions are minimized by different optimization strategies. Modeling the problem as an energy minimization process, these methods aim to produce a desirable segmentation by achieving global optimization. However, these methods only provide a characterization of each cut rather than the whole regions and they might be computationally inefficient for many practical applications. In most cases, it is very difficult to find the globally optimal solution for cut based energy functions. Since a single optimal partition of an image is not easy to obtain, it makes a possibility of finding different level-based explanations of an image. From this aspect, there are works [11-12] tackling the image segmentation as a hierarchical bottom-up problem.

Another classical category of segmentation algorithms is based on the similarity among the pixels within a region, namely region-based segmentation. A lot of work has investigated the use of primitive regions as a preprocessing step for image segmentation [13-15]. The advantages are twofold: first, regions carry on more information in describing the nature of objects. Second, the number of primitive regions is much fewer than that of pixels in an image and thus largely speeds up the region merging process. Starting from a set of primitive regions, the segmentation is conducted by progressively merging the similar neighboring regions according to a certain predicate. Although the segmentation is obtained by making local decisions, some techniques have been proved to be very efficient [16-21]. In region merging techniques, the goal is to merge regions that satisfy a certain homogeneity criterion. In previous works, there are region merging algorithms based on statistical properties [16, 22, 23], graph properties [17-19, 24-25] or spatio-temporal similarity [21]. However, most region merging algorithms are hardly proven to preserve the global properties.

In this paper, we implement the segmentation algorithm in a region merging style for its merit of efficiency, where similar neighboring regions are iteratively merged according to a novel merging predicate. The proposed predicate can be interpreted as a combination of the consistency measure and the similarity measure. As stated above, homogeneity criteria (cues) are essential to the region merging process. However, in this paper we do not focus on how to find good cues. Instead, we model the cues as a function of random variables. Evaluation of the homogeneity of regions follows the principle of Gestalt theory [26-27], which suggests a preference to having consistent elements in the same data set. Unlike some traditional methods which measure the data consistency with a certain threshold [20], the function we use is associated to a binary value which is defined by the Sequential Probability Ratio Test (SPRT) [28]. More specifically, the extent of consistency is measured by two hypotheses, i.e. null hypothesis $H_0$ "the tested data are inconsistent" and an alternative hypothesis $H_1$ "the tested data are consistent". The similarity measure describes the affinity between two neighboring regions. In each iteration, a region finds its closest neighbor at the lowest value according to some cost function. This casts as that of finding in a spatial-time search space an optimal path that most satisfies the data consistency. If taking the image segmentation as a labeling problem [1], the label of a region will transit to the one of its closest neighbor as iteration goes on. As a result, a merging will happen when two neighboring regions exchange their labels. We can prove that this merging scheme satisfies the maximum likelihood criterion, by which a primitive region will be merged into its most similar group at the lowest cost in the final segmentation.

A good segmentation algorithm should preserve certain global properties according to the perceptual cues [16, 22-23]. This leads to another essential problem in a region merging algorithm: the order that is followed to perform the region merging. Since the merging process is inherently local, most existing algorithms have difficulties to possess some global optimality. However, we demonstrate that the proposed algorithm holds certain global properties, i.e. being neither over-merged nor under-merged, using the defined merging predicate. In addition, to speed up the region merging process, we introduce the structure of *Nearest Neighbor Graph* to accelerate the proposed algorithm in searching the merging candidates. The experimental results indicate the efficiency of the acceleration algorithm.

The rest of this paper is organized as follows. In Section II, a novel region merging predicate is defined. The consistency test of neighboring regions is defined in Section III by using SPRT. In Section IV, we present the dynamic region merging algorithm and prove some global properties it holds for the segmentation. In Section V, an accelerated algorithm for the merging process is presented and analyzed. In Section VI, we show the segmentation results on real images. Section VII concludes the paper.

## II. The Region Merging Predicate

Automatic image segmentation can be phrased as an inference problem [1]. For example, we might observe the colors in an image, which are caused by some unknown principles. In the context of image segmentation, the observation of an image is given but the partition is unknown. In this respect, it is possible to formulate the inference problem as finding some representation of the pixels of an image, such as the label that each pixel is assigned. With these labels, an image is partitioned into a meaningful collection of regions and objects. The Gestalt laws in psychology [26-27] have established some fundamental principles for this inference problem. For example, they imply some well-defined perceptual formulations for image segmentation, such as homogeneous, continuity and similarity. In the family of region merging techniques, some methods have used statistical similarity tests [16, 29] to decide the merging of regions, where a predicate is defined for making local decisions. These are good examples of considering the homogeneity characteristics within a region, from which we can see that an essential attribute for region merging is the consistency of data elements in the same region. In other words, if neighboring regions share a common consistency property, they should belong to the same group. However, most of the existing region merging algorithms cannot guarantee a globally optimal solution of the merging result; in other words, the region merging output is either over-segmented or under-segmented. In this section, we propose a novel predicate which leads to certain global properties for the segmentation result.

The proposed predicate is based on measuring the dissimilarity between pixels along the boundary of two regions. For the convenience of illustrating the whole framework, we use the definition of *region adjacency graph* (RAG) [30] to represent an image. Let $G = (V, E)$ be an undirected graph, where $v_i \in V$ is a set of nodes corresponding to image elements (e.g. super-pixels or regions). $E$ is a set of edges connecting the pairs of neighboring nodes. In other words, between two nodes there exists an edge if the nodes are adjacent. Each edge $(v_i,v_j) \in E$ has a corresponding weight $w((v_i,v_j))$ to measure the dissimilarity of the two nodes connected by that edge. In the context of region merging, a region is represented by a component $R \subseteq V$. We obtain the dissimilarity between two neighboring regions $R_1, R_2 \subseteq V$ as the minimum weight edge connecting them. That is,

$$S(R_1, R_2) = \min_{v_i \in R_1, v_j \in R_2, (v_i, v_j) \in E} w((v_i, v_j)) \qquad (1)$$

The graph structure of an example partition is shown in Fig. 1, where the image has 7 partitioned regions and its RAG is accordingly shown on the right. The advantage of RAG is that it can provide a "spatial view" of the image.

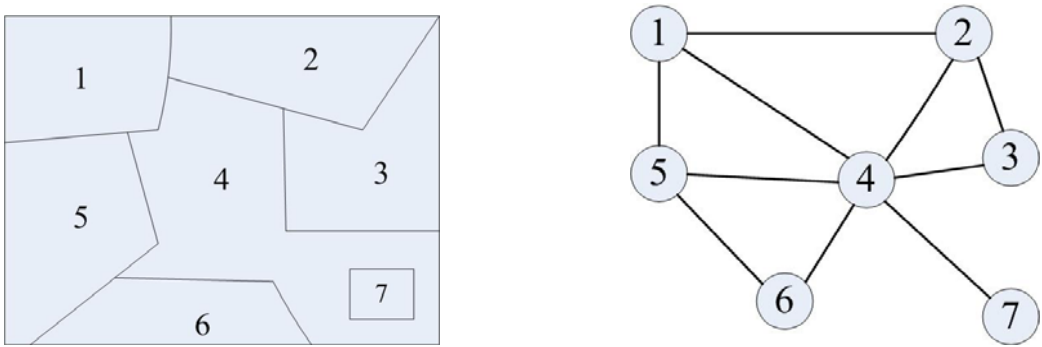

**Figure 1.** An example of region partition and the corresponding region adjacency graph (RAG).

Since the merging predicate will decide whether there is an evidence of merging between the pair of regions, it involves two aspects: a dissimilarity measure which is used to determine the candidate region for merging, and the consistency property which checks if the regions are homogenous. We define the following region merging predicate *P*:

$$P(R_1, R_2) = \begin{cases} true & \text{if (a) } S(R_1, R_2) = \min_{R_i \in \Omega_1} S(R_1, R_i) = \min_{R_j \in \Omega_2} S(R_2, R_j); \text{ and} \\ & \quad\quad (b) R_1 \text{ and } R_2 \text{ are consistent} \\ false & \text{otherwise} \end{cases} \quad (2)$$

where $\Omega_1$ and $\Omega_2$ are the neighborhood sets of $R_1$ and $R_2$, respectively. The merging predicate on regions $R_1$ and $R_2$ could thus be "merge $R_1$ and $R_2$ if and only if they are the most similar neighbors in each other's neighborhood and follow the principle of consistency." The condition (a) is stronger than that of only requiring the connecting edge between $R_1$ and $R_2$ to be the minimum one in either of the neighborhood. This leads to an interesting property of the proposed region merging algorithm, i.e., it is not influenced by the starting point of merging. We shall see hereafter that such a condition uniquely decides the pairs of regions to be merged in a given merging level. Moreover, in Section V we will prove that there is always at least one pair of regions which satisfies condition (a). Clearly, without condition (b), all the regions will merge into one big region at the end of region merging process. Therefore, condition (b) acts as a stopping criterion. Fig. 2 illustrates an example when the predicate $P$ between regions $R_1$ and $R_2$ is true.

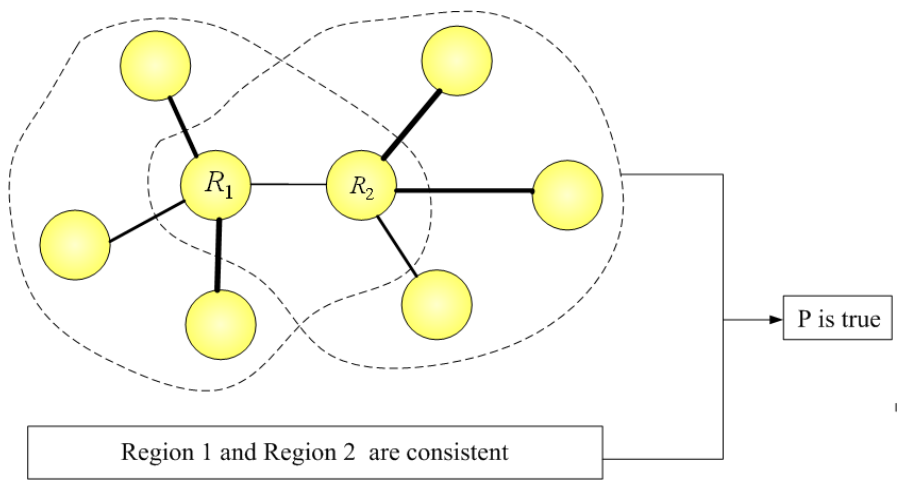

**Figure 2.** An example that the predicate $P$ between $R_1$ and $R_2$ is true. The thickness of lines indicates the weights of the edges. The most similar pair of regions is connected by an edge with the minimum weight.

According to the definition of $P$, the test of consistency is based on the visual cues extracted from the image data. In the next section, the SPRT method is introduced for a reliable decision on the consistency of regions.

## III. Consistency Test of Cues

In order to obtain the homogenous regions in region merging, the proposed predicate *P* in Eq. (2) checks the consistency of regions. Region information is usually presented by the cues extracted from the observed data. The choice of cues can be intensity, color, texture and so on. If we view the cue as a random variable, the distribution of the cue depends on the consistency of pairs of regions. In this paper, we formulate the evaluation of the region consistency as a sequential test process. Suppose parameter $\theta$ is related to the distribution of random cues *x*. More specifically, we gather information of parameter $\theta$ by observing random variables in successive steps. Since every sample of the cues carries statistical information on parameter $\theta$, we may collect the information at the end of observation. This is one of the interesting problems studied in sequential analysis, where $\theta$ is called a hypothesis. In the context of region merging, two hypotheses are involved in the evaluation task: a pair of regions is "consistent", and is "inconsistent", which are denoted by a null hypothesis $H_0$: $\theta = \theta_0$ against an alternative hypothesis $H_1$: $\theta = \theta_1$, respectively. The property of the hypotheses is a hidden state that is not directly observable, but is statistically linked to the observable cues. To decide whether or not a pair of regions belongs to the same group, we look for the solution of its hypothesis test.

An efficient and popular procedure for integrating the statistical evidence is the *Sequential Probability Ratio Test* (SPRT) which was proposed by Wald [28]. SPRT shows that the solution to the hypothesis can be found by making the smallest number of observations and satisfying the bounds on the predefined probabilities of two errors. SPRT is purely sequential in nature, i.e. continuing sampling on the instances of a random variable will eventually lead to a reliable inference about parameter $\theta$.

The application of SPRT to the consistency test of cues is described as follows. We observe the distribution of random cues *x* in a sequence until a likelihood ratio $\delta$ goes out of the interval (*B,A*) for the first time by a random walk, where the real numbers *A* and *B* satisfy *B<0<A*. The sequence of successive likelihood ratio $\delta_i$ is:

$$\delta_i = \log \frac{P_1(x_i | \theta_1)}{P_0(x_i | \theta_0)}, \quad i = 1, 2, ..., N \tag{3}$$

where $P_0(x|\theta_0)$ and $P_1(x|\theta_1)$ are the distributions of visual cues. $P_0(x|\theta_0)$ and $P_1(x|\theta_1)$ should be different so as to make a convincing decision. We use the Gaussian distribution model to approximate the cue distributions.

The two conditional probabilities are given as follows:

$$\begin{cases} P_0(x|\theta_0) = 1 - \lambda_1 \exp(-(I_b - I_a) S_I^{-1} (I_b - I_a)) \\ P_1(x|\theta_1) = 1 - \lambda_2 \exp(-(I_b - I_{a+b}) S_I^{-1} (I_b - I_{a+b})) \end{cases} \quad (4)$$

where $I_a$, $I_b$ are the average color of sampled data in region $a$, $b$ respectively, and $I_{a+b}$ is the average value of samples' union. $S_I$ is the covariance matrix of the regions, and $\lambda_1$, $\lambda_2$ are scalar parameters. If each test is independent, the composition of the likelihood ratios is the sum of the individual $\delta_i$:

$$\delta = \sum_{i=1}^{N} \delta_i \quad (5)$$

where $N$ is the first integer for which $\delta \geq A$ or $\delta \leq B$. We can see that the solution to the hypothesis is decided by the relationship between $\delta$ and an upper and lower limits, denoted by $A$ and $B$, respectively. If $\delta$ goes out of one of these limits, the hypothesis is made and thus the test stops. Otherwise, the test is carried on with a new random sampling.

Intuitively, the likelihood ratio $\delta$ is positive and high when the terminal decision is in favor of $H_1$, while it is negative and low when the situation is reversed. In the SPRT theory [28], Wald recommended implementing the test with a practical approximation: $A=\log(1-\beta)/\alpha$ and $B=\log\beta/(1-\alpha)$, where $\alpha$ and $\beta$ are probabilities of the decision error given by:

$$\alpha = \Pr\{\text{Rejecting } H_0 \text{ when } H_0 \text{ is true}\}$$
$$\beta = \Pr\{\text{Accepting } H_0 \text{ when } H_1 \text{ is true}\}$$

The selection of $\alpha$ and $\beta$ affects the region merging quality. Intuitively, as error rates decrease, the region merging quality grows. However, at the same time the computational effort will increase accordingly. In our implementation, both $\alpha$ and $\beta$ are set as a fixed value 0.05. If consider the probabilities of the decision error, the likelihood ratio becomes:

$$\delta_i = \log \frac{P_1(x_i|\theta_1) \cdot (1-\beta)}{P_0(x_i|\theta_0)(1-\alpha)} \quad (6)$$

For the uncertainty of the (worst-case) number of tests in SPRT, a truncated SPRT [28] is used here by presetting an upper bound $N_0$ on the number of tests. In Wald's theory, the expected number of tests is given by:

$$\begin{cases} E\{n|\theta_0\} = [A\alpha + B(1-\alpha)]/\eta_0 \\ E\{n|\theta_1\} = [A(1-\beta) + B\beta]/\eta_1 \end{cases} \quad (7)$$

where $\eta_0$, $\eta_1$ are the conditional expected number of trails from a single test: $\eta_0=E\{\delta|\theta_0\}$ and $\eta_1=E\{\delta|\theta_1\}$. We set $N_0$ to be a constant which is greater than $\max\{E\{\delta|\theta_0\}, E\{\delta|\theta_1\}\}$. The proposed SPRT based consistency test of cues is summarized in Table 1.

Table 1. **Algorithm 1**: consistency test of cues

**Preset** $\lambda_1$;
**Let** $\lambda_2=1$, $\alpha=0.05$, $\beta=0.05$;
**Compute parameters:**
$N_0$: be a constant greater than $\max\{E\{\delta|\theta_0\}, E\{\delta|\theta_1\}\}$;
$A=\log(1-\beta)/\alpha$, $B=\log\beta/(1-\alpha)$;
$P_0(x|\theta_0)$, $P_1(x|\theta_1)$ are computed using Eq. (4).

**Input**: a pair of neighboring regions.
**Output**: the decision $D$ that the two regions are "consistent" ($D=1$) or "inconsistent" ($D=0$).

1. Set evidence accumulator $\delta$ and the trials counter $n$ to be 0.
2. Randomly choose $m$ pixels in each of the pair of regions, where $m$ equals the half size of the region.
3. Calculate the distributions of visual cues $x$ using Eq. (4) based on these pixels.
4. Update the evidence accumulator $\delta = \delta + \log\frac{P_1(x|\theta_1)(1-\beta)}{P_0(x|\theta_0)(1-\alpha)}$.
5. If $n \leq N_0$
    If $\delta \geq A$, return $D=1$ (consistent)
    If $\delta \leq B$, return $D=0$ (inconsistent)
   If $n > N_0$
    If $\delta \geq 0$, return $D=1$ (consistent)
    If $\delta < 0$, return $D=0$ (inconsistent)
6. Go back to step 2.

## IV. Dynamic Region Merging (DRM)

### IV-A. The dynamic region merging algorithm

In this section, we explain the proposed region merging algorithm as a dynamic region merging (DRM) process, which is proposed to minimize an objective function with the merging predicate $P$ defined in Eq. (2). As mentioned in Section I, the proposed DRM algorithm is started from a set of over-segmented regions. This is because a small region can provide more stable statistical information than a single pixel, and using regions for merging can improve a lot the computational efficiency. For simplicity and in order to validate the effectiveness of the proposed DRM algorithm, we use the watershed algorithm [31] (with some modification) to obtain the initially over-segmented regions (please refer to Section VI-A for more information), yet using a more sophisticated initial segmentation algorithm (e.g. mean-shift [32]) may lead better final segmentation results.

Given an over-segmented image, there are many regions to be merged for a meaningful segmentation. By taking the region merging as a labeling problem, the goal is to assign each region a label such that regions belong to the same object will have the same label. There are two critical labels for a region $R_i$: the initial label $l_i^0$, which is decided by the initial segmentation, and the final label $l_i^n$, which is assigned to the region when the merging process stops. In our problem, the final label $l_i^n$ for a given region is not unique, which means that the same initialization $l_i^0$ could lead to different solutions. This uncertainty mainly comes from the process of SPRT with a given decision error. However, it can be guaranteed that all the solutions satisfy the merging predicate $P$ defined in Eq. (2). In the process of region merging, the label of each region is sequentially transited from the initial one to the final one, which is denoted as a sequence ($l_i^0$, $l_i^1$,…, $l_i^n$).

To find an optimal sequence of merges which produce a union of optimal labeling for all regions, the minimization of a certain objective function $F$ is required. According to predicate $P$, the transition of a region label to another label corresponds to a minimum edge weight connects the two regions. In this case, a sequence of transitions will be defined on a set of local minimum weights, i.e., in each transition the edge weight between the pair of merged regions should be the minimum one in the neighborhood. As a result, the objective function $F$ used in this work is defined as the measure of transition costs in the space of partitions. In other words, as the whole image is a union of all regions, $F$ is the sum of transition costs over all regions. That is:

$$F = \sum_{R_i} F_i \qquad (8)$$

where $F_i$ is the transition costs of one region $R_i$ in the initial segmentation. Minimizing $F$ in Eq. (8) is a combinatorial optimization problem and finding its global solutions is in general a hard task. Since the exhaustive search in the solution space is impossible, an efficient approximation method is desired. The solution adopted here is based on the stepwise minimization of $F$, where the original problem is broken down into several sub-problems by using the dynamic programming (*DP*) technique [33].

The *DP* is widely used to find the (near) optimal solution of many computer vision problems. The principle of *DP* is to solve a problem by studying a collection of sub-problems. Indeed, there have been some works in image segmentation that benefit from this efficient optimization technique, such as *DP* snake [34,35]. In the proposed DRM algorithm, we apply *DP* on discrete regions instead of line segments [34, 35]. The minimization problem for region $R_i$ starting at labeling $l_i^0$ is defined as:

$$\begin{aligned}
\min F_i(l_i^0, \cdots, l_i^n) &= \min F_i(l_i^0, l_i^{n-1}) + d_{n-1,n} \\
&= \min F_i(l_i^0, l_i^{n-2}) + d_{n-2,n-1} + d_{n-1,n} \\
&= \cdots \\
&= \sum_{k=0}^{n-1} d_{k,k+1}
\end{aligned} \quad (9)$$

where $F_i(l_i^0, l_i^1, \ldots, l_i^n)$ is the transition cost from $l_i^0$ to $l_i^n$, $d_{k,k+1}$ is the minimum edge weight between the regions with labeling $l_i^k$ and $l_i^{k+1}$, respectively. In conjunction with Eq. (1), we have

$$d_{k,k+1} = \min_{R_{k+1} \in \Omega_k} S(R_k, R_{k+1}) \quad (10)$$

The overall path length from $l_i^0$ to $l_i^n$ is the sum of minimum edges $d_{k,k+1}$ for each node in that path. This problem reduces to a search for a shortest path problem, whose solution can be found at a finite set of steps by the Dijkstra's algorithm in polynomial time. At this point, a minimization process of object function $F$ is exactly described by the predicate *P* defined in Eq. (2), where *P* is true if the nodes are connected by the edge with the minimum weight in their neighborhood. It means that the closest neighbors will be assigned to the same label, which is the cause of the merging.

In Fig. 3, an example process of region merging is shown by embedding it into a 3D graph. Between two adjacent layers there is a transition which indicates the costs of a path. Clearly, this is also a process of label transitions. The neighborhood of the highlighted region (in red) is denoted as the black nodes in the graph and the closest neighbor is denoted as the red nodes. The directed connections with the lowest cost between adjacent layers are made (shown as blue arrows). Note that the connectivity between regions in the same layer is represented by the RAG, which is not explicitly shown in Fig. 3.

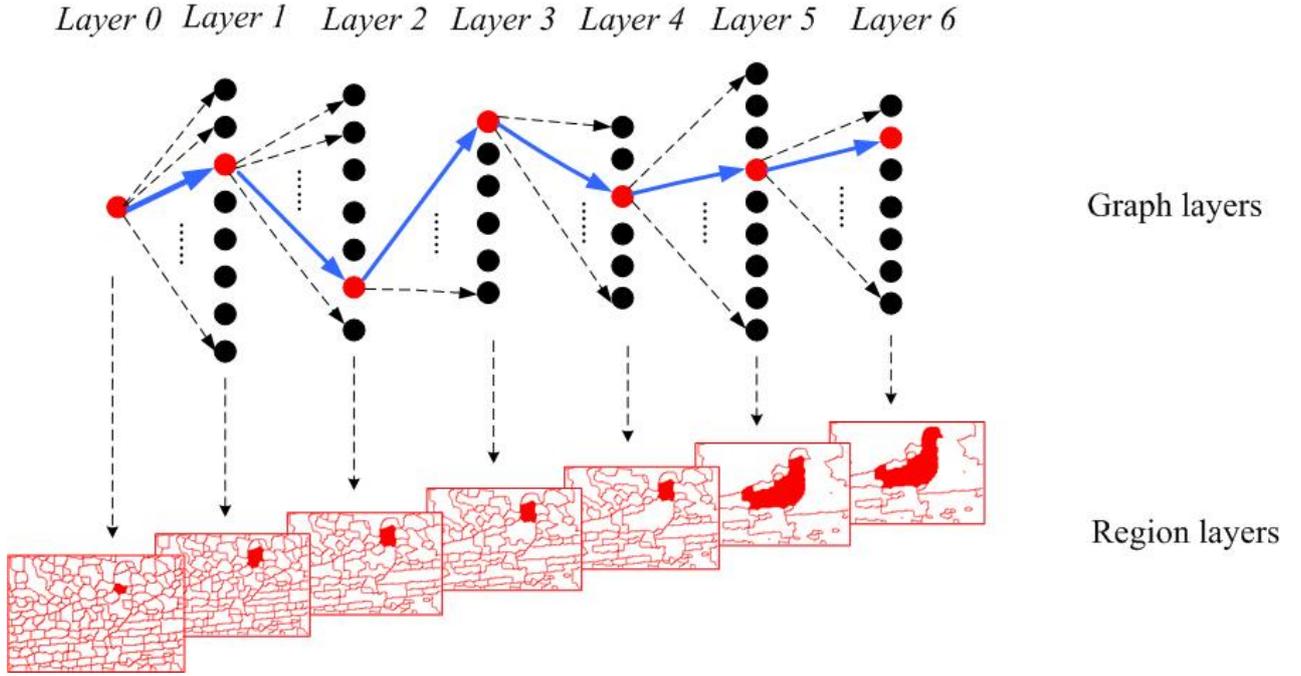

**Figure 3.** The dynamic region merging process as a shortest path in a layered graph. The upper row shows the label transitions of a graph node. The lower row shows the corresponding image regions of each label layer. Starting from layer 0, the highlighted region (in red) obtains a new label from its closest neighbor (in red). If the region is merged with its neighbor, they will be assigned to the same label. The shortest path is shown as the group of the directed edges (in blue).

The proposed DRM algorithm is summarized in Table 2. Minimization of the objective function follows the principle of *DP*, which is exactly expressed in terms of predicate *P*. In the *k*-th decision step, a merging occurs when two neighboring regions are connected by the minimum weight edge in each other's neighborhood. With the observation of this characteristic, we will propose an accelerated region merging algorithm in Section V.

**Table 2. Algorithm 2**: segmentation by dynamic region merging

| |
|---|
| **Input**: the initially over segmented image $S_0$. |
| **Output**: region merging result. |
| 1. Set $i=0$. |
| 2. For each region in segmentation $S_i$, use **Algorithm 1** to check the value of predicate *P* with respect to its neighboring regions. |
| 3. Merge the pairs of neighboring regions whose predicate *P* is true, such that segmentation $S_{i+1}$ is constructed. |
| 4. Go back to step 2 until $S_{i+1} = S_i$. |
| 5. Return $S_i$. |

## IV-B. Properties of the proposed DRM algorithm

Although the proposed DRM scheme is conducted in a greedy style, some global properties of the segmentation can be obtained. More specifically, it can be proved that the proposed DRM algorithm produces a segmentation $S$ which is neither over-merged nor under-merged according to the proposed predicate $P$. Similar to the definitions of over-segmentation and under-segmentation in [18], we define the concepts of over-merged segmentation and under-merged segmentation as below.

*Definition 1*. *Under-merged segmentation*. A segmentation $S$ is under-merged if it contains some pair of regions for each there is an evidence of a merging between the regions.

*Definition 2*. *Over-merged segmentation*. A segmentation $S$ is over-merged, if there is another segmentation $S_r$ which is not under-merged and each region of $S_r$ is contained in some component of $S$. Or saying, $S_r$ can be obtained by splitting one or more regions of $S$. We call that $S_r$ is a refinement of $S$.

*Definition 3. The evidence of boundary*. There is an evidence of boundary between a pair of regions $R_1$ and $R_2$ if the predicate $P$ is false because the inconsistency property between them does not hold, with respect to Eq. (2).

*Lemma 1*. If two adjacent regions are not merged for an evidence of a boundary between them in the $k$-th iteration, they will be in different regions in the final segmentation. Denote by $R_i^k$ and $R_j^k$ two neighboring regions in the $k$-th iteration. Then $R_i = R_i^k$ and $R_j = R_j^k$, where $R_i$ is the region whose label is $L_i$ and $R_j$ is the region whose label is $L_j$ in the final segmentation $S$.

*Proof*: If there is an evidence of a boundary between $R_i^k$ and $R_j^k$, according to **Definition** 3, we have $S(R_i, R_j) = \min_{a \in \Omega_i} S(R_i, R_a) = \min_{b \in \Omega_j} S(R_j, R_b)$, where $\Omega_i$ and $\Omega_j$ are neighborhood of $R_i^k$ and $R_j^k$. Since $R_i^k$ and $R_j^k$ are not merged, they will not be merged with any other regions in the remaining steps before they are merged with each other. So we have $R_i = R_i^k$ and $R_j = R_j^k$. ∎

**Lemma** 1 holds because our merging predicate relies on comparing the minimum weight edge between regions. We can prove that with this measure, some global properties of the segmentation can be easily obtained.

*Theorem 1*. The segmentation $S$ by **Algorithm** 2 is not under-merged according to **Definition** 1.

*Proof*. If $S$ is under-merged, there must be some pair of regions $R_i^k$ and $R_j^k$ that do not cause a merge in the merging process. Therefore, the evidence of a boundary does not hold for $R_i^k$ and $R_j^k$. According to **Lemma** 1, if $R_i^k$ and $R_j^k$ are not merged, $R_i = R_i^k$ and $R_j = R_j^k$. This implies that **Algorithm 2** does not merge $R_i^k$ and $R_j^k$. The evidence of a boundary holds for them, which is a contradiction. ∎

*Theorem 2.* The segmentation $S$ by **Algorithm** 2 is not over-merged according to **Definition** 2.

*Proof.* If $S$ is over-merged, there must be a proper refinement $T$ that is not under-merged. Let a region $C \in S$, and there are two adjacent regions $A \subset C$, $B \subset C$, such that $A$ and $B$ satisfy the refinement $T$. According to **Algorithm** 2, $A$ and $B$ will not merge with any other regions in $C$ before they merge with each other. Then $C$ does not contain $A$ and $B$, which is a contradiction. ∎

## V. Algorithm Acceleration by Nearest Neighbor Graph (NNG)

The DRM process presented in **Algorithm** 2 depends on adjacency relationships between regions. At each merging step, the edge with the minimum weight in a certain neighborhood is required. This requires a scan of the whole graph by which the relations between neighboring regions are identified. The edge weights and nodes of RAG are calculated and stored for each graph layer. Since the positions of the edges are unknown, the linear search for these nodes and edges requires $O(||E||)$ time. After each merging, at least one of the edges must be removed from RAG, the positions and edge weights are updated. Then a new linear search is performed for constructing the next graph layer. If the number of regions to be merged is very large, the total computational cost in the proposed DRM algorithm will be very high.

Based on the observation that only a small portion of RAG edges counts for the merging process, we can find an algorithm for accelerating the region merging process. The implement of the algorithm relies on the structure of nearest neighbor graph (NNG), which is defined as follows. For a given RAG, where $G = <V, E>$, the NNG is a directed graph $G_m = <V_m, E_m>$, where $V_m = V$. If we define a symmetric dissimilarity function $S$ to measure the edge weights, the directed edge is defined as:

$$E_m = \{(v_i, v_j) \mid w((v_i, v_j)) = \min S(v_i, v_k), (v_i, v_k) \in E\}$$

Fig. 4 shows an example of NNG with respect to the RAG in Fig. 1. From the definition we can see that the out-degree of each node equals to one and the number of edges in an NNG is $||V||$. The edge starting at a node is pointing toward its most similar neighbor. For a sequence of graph nodes, if the starting and ending nodes coincide, we call it a cycle (see Fig. 4).

It is easy to verify that the NNG has the following properties [36]:

(1) Along any directed edge in NNG, the weights are non-increasing.

(2) The maximum length of a cycle is two.

(3) NNG contains at least one cycle.

(4) The maximum number of cycles is $\lfloor \|V\|/2 \rfloor$.

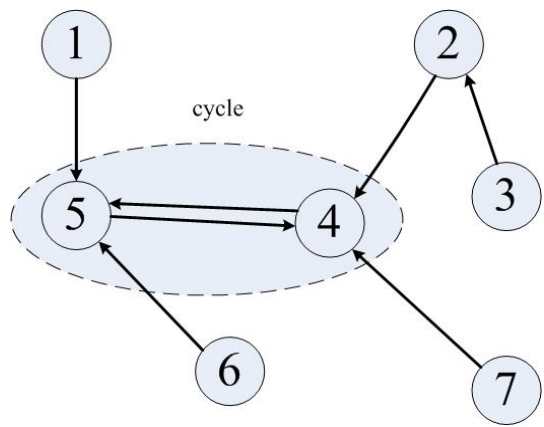

**Figure 4.** A possible NNG of the RAG in Fig. 1 and a cycle in the NNG.

From the definition of merging predicate in Eq.(2), we see that when the predicate value between two regions is true, there is exactly a cycle between them. This demonstrates the stop criterion for the proposed DRM algorithm, i.e., if there is no cycle in the NNG, the region merging will stop. In other words, before the process stops, we can always have at least one pair of regions to merge according to the above property (3). This suggests that we can keep the NNG cycles during the region merging process instead of searching over the whole RAG. The original RAG is constructed from the over-segmented image, and the NNG is formed by searching for the most similar neighbors of each graph node. The NNG cycles are identified by a scan of the NNG. The accelerated region merging process is described in Table 3.

Now we have the following theorem to demonstrate that the regions merging process will go on until the predicate is false for all pairs of regions in the image.

*Theorem 3*. In the DRM algorithm, there is at least one pair of regions to be merged in each iteration before the stopping criterion is satisfied.

*Proof*. According to the property (3) of the NNG, there is always at least one cycle in the graph (the number of graph nodes should be greater than 1). Therefore, the region merging process will continue until the condition (b) in Eq.(2) is not satisfied. ∎

Table 3. **Algorithm** 3: Accelerating the dynamic region merging process.

| |
|---|
| **Input**: the initial RAG and NNG of the image. <br> **Output**: The region merging result (in the form of RAG). |
| 1. Set $i=0$. <br> 2. For the NNG in the $i$-th graph layer, find the minimum weight edge of the RAG using the cycles. <br> 3. Use **Algorithm 1** to check the value of the predicate $P$. If $P$ is between the cycle is true, merge the corresponding pair of regions. <br> 4. Update the RAG, NNG and the cycles. <br> 5. Set $i=i+1$. <br> 6. Go back to step 2 until no cycle can be found. <br> 7. Return RAG. |

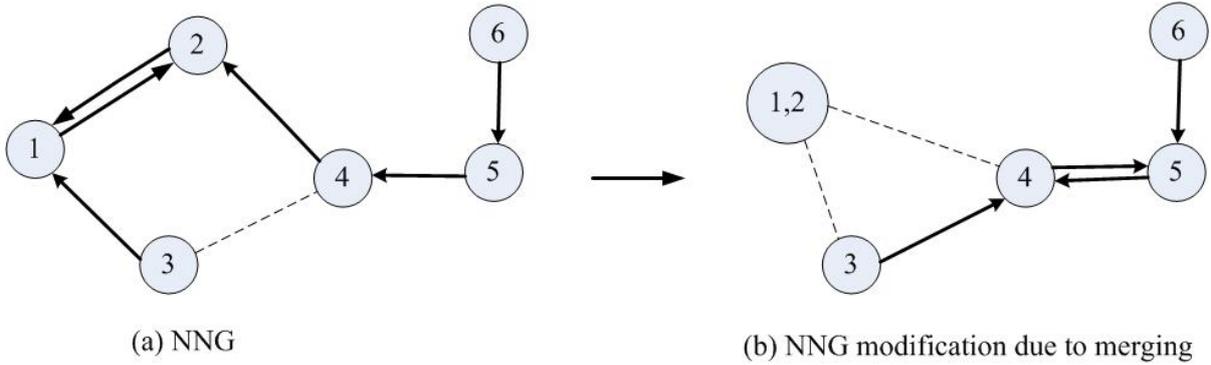

**Figure 5.** An example of NNG modification. Dotted lines represent the RAG edges, while directed lines represent NNG edges.

If only use RAG in the merging process, update of the RAG requires $O(\|V\|)$ searching time. The computation is usually expensive due to a large number of nodes in RAG. In Algorithm 3, after the nodes of a cycle are merged, the weights of the neighboring RAG and the structure of NNG are modified. There is an observation that the new cycles can only form in the second order neighborhood of the merged nodes, which is illustrated in Fig. 5. After merging nodes 1 and 2, the closest neighbor of node 3 becomes node 4, and that of node 4 becomes node 5. A new cycle is formed between nodes 4 and 5, and by no means to be between nodes 5 and 6. In such a way, the causes of new NNG cycles can only be detected in the second order neighborhood of merged nodes. Hence, the computational effort for updating the NNG at each merge of region pair depends on the distribution of the second order neighborhood size in the RAG. The computation time for a merge of NNG cycle is $O(\gamma^{(2)}+1)$, where $\gamma^{(2)}$ is the size of the second order neighborhood of the new node. In most cases, $\gamma^{(2)}$ is far less than $\|V\|$, which indicates the reduction of computation time by the accelerating algorithm.

## VI. Experimental Results and Discussions

In this section, we evaluate the proposed DRM algorithm on the Berkeley Segmentation Dataset[2] (BSDS), which contains 100 test images with 5-10 human segmentations on each one of them as the ground-truth data. In Section VI-A, we test the DRM algorithm with several representative examples. In Section VI-B, we compare the well known mean-shift algorithm [32] and the graph-based region merging method [18] with the DRM algorithm. In Section VI-C, the performance of the accelerated DRM algorithm is evaluated. In Section VI-D, we discuss the DRM method and its potential extensions.

### VI-A. Analysis on DRM by representative examples

The proposed DRM algorithm starts from an initially over-segmented image. For simplicity, we use the watershed algorithm to obtain the initial segmentation. Certainly, a more sophisticated initial segmentation method, such as the mean-shift algorithm, may lead to a better final segmentation result. Since the standard watershed algorithm is very sensitive to noise and hence leads to severe over-segmentation (see Fig 6(b) for an example), to reduce noise and trivial structures we apply median filtering on the gradient image before using the watershed algorithm. Fig. 6(c) shows the result of the modified watershed segmentation, where the over-segmentation is reduced a lot while preserving the desired the object boundaries.

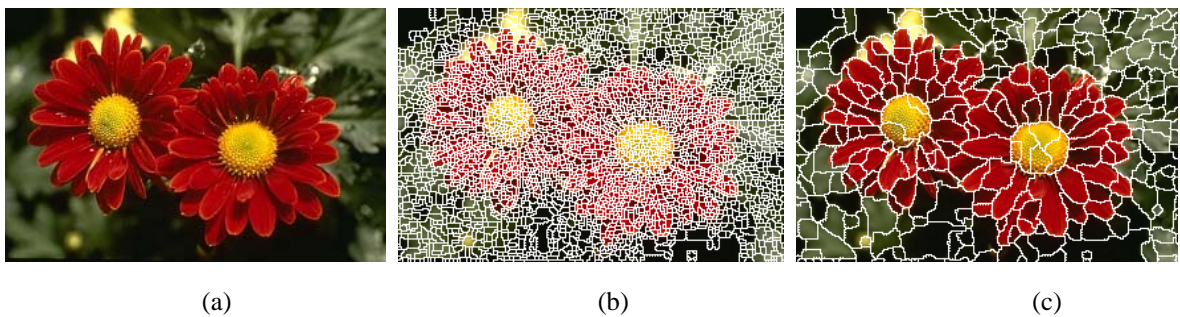

(a)                   (b)                   (c)

**Figure 6.** (a) Original image; (b) initial segmentation by standard watershed algorithm; and (c) initial segmentation by modified watershed algorithm with median filtering on the gradient image.

---

[2] http://www.cs.berkeley.edu/projects/vision/grouping/segbench/

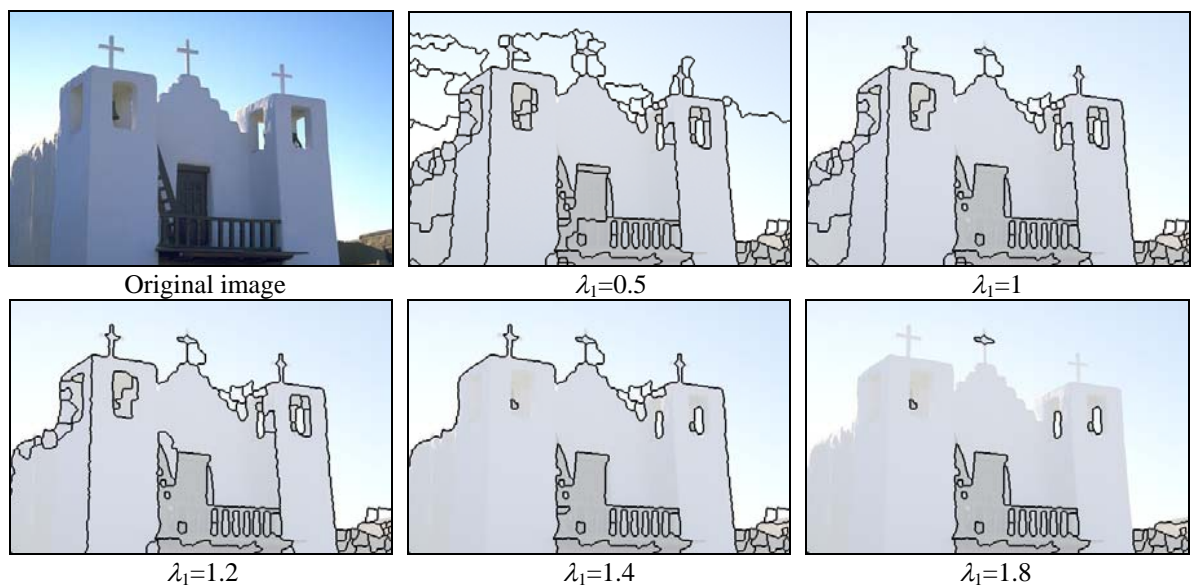

**Figure 7**. Segmentation results with $\lambda_1$=0.5, 1, 1.2, 1.4, 1.8, respectively.

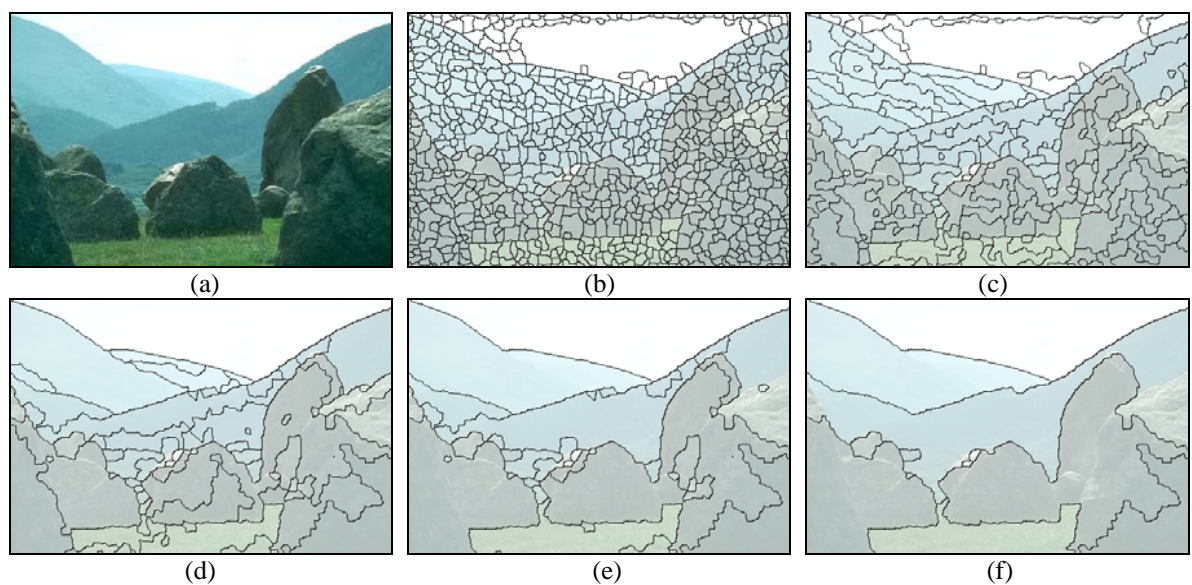

**Figure 8**. Region merging process. (a) the original image; (b) the result of initial watershed segmentation; (c)-(f) the merging results in different stages.

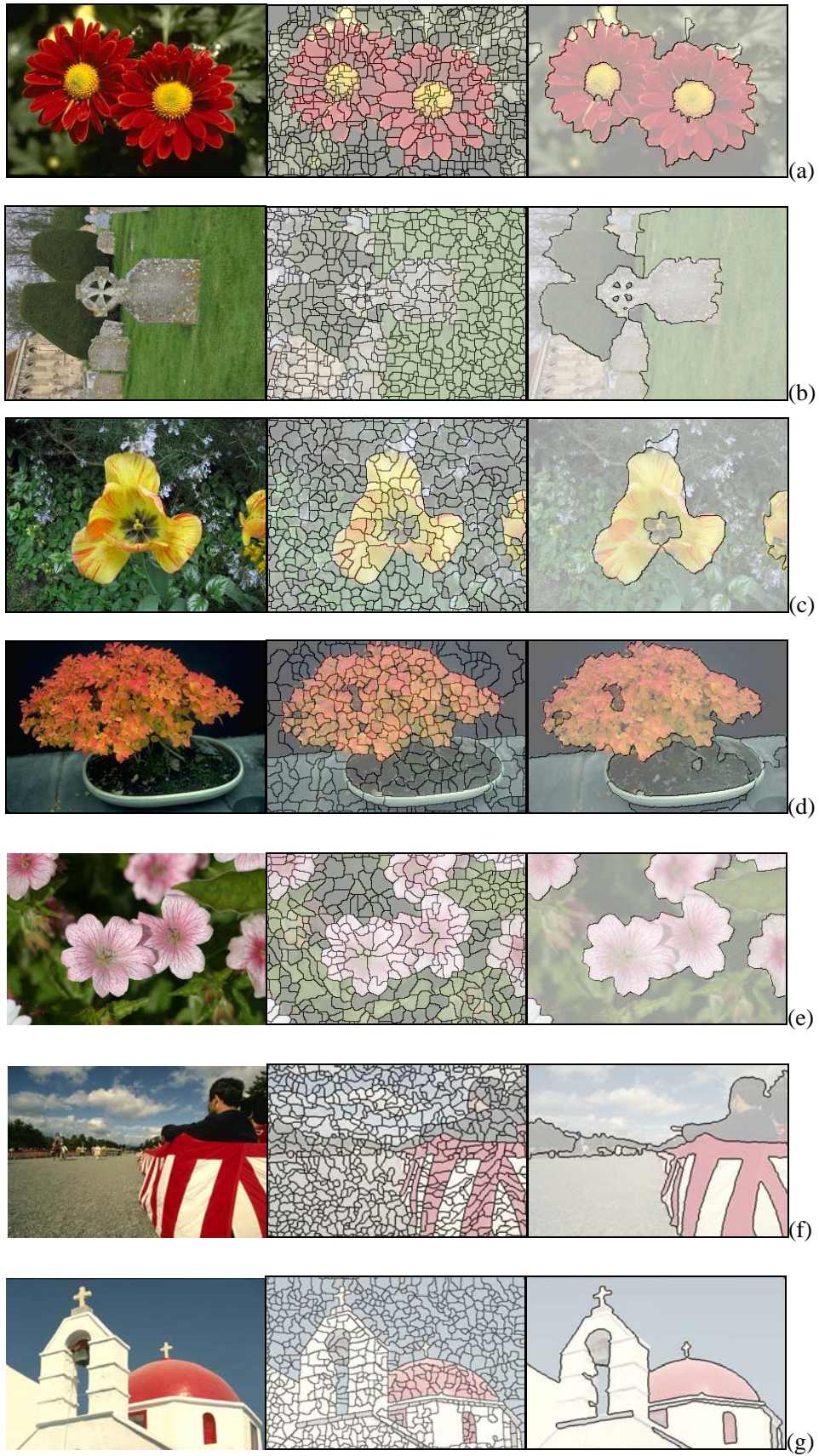

**Figure 9.** Segmentation results by the proposed algorithm. From left to right, the first column shows the original images. The second column shows the over-segmentation produced by watershed algorithm. The third column shows our segmentation results.

As stated in Section III, there is only one set of parameters in the proposed algorithm, i.e. the values of $\lambda_1$, $\lambda_2$ in the cue distribution functions (refer to Eq. (4)). These parameters control the degree of consistency between two regions, and hence decide when to terminate the region merging. In practice, we set $\lambda_2$ to be a constant value 1, and hence there is only one parameter $\lambda_1$ to be set. By experiment experience, we set $\lambda_1$ between 0.5 and 5. Fig. 7 shows the segmentation results of an example image with different values of $\lambda_1$. When $\lambda_1$ is large, the algorithm is more likely to take two neighboring regions as consistency. Therefore it may produce over-merged regions (see the result where $\lambda_1=1.8$). When $\lambda_1$ is small, the case is inverse and as a result produces under-merged regions (see the results where $\lambda_1=0.5, 1$). To clearly show the region merging process, we give an example in Fig.8, in which primitive regions (in Fig. 8(b)) are merged iteratively until the stop criterion is satisfied (in Fig. 8(f)).

We illustrate the proposed DRM algorithm using more example images in Fig. 9. It is clear that neighboring regions with coherent colors are merged into one, whereas the boundaries are well located on the reasonable places. Some of the large regions have substantial variations inside (Figs. 9 (c), (e), (f)), however, with relatively slow changes of colors along the boundaries. This indicates that the DRM algorithm can tolerate some variations for grouping meaningful regions in an image.

**VI-B. Comparison with state-of-the-art methods**

In this section we compare the segmentation results of DRM with the mean-shift algorithm [32] and the graph-based algorithm [18], which represent state-of-the-art of automatic image segmentation. Mean-shift performs the segmentation by clustering the data into several disjoint groups. The cluster centers are computed by first defining a spherical window, then shifting the window to the mean of the data points with a radius *r*. This shifting process repeats until convergence. The segmented image is constructed using the cluster labels. Mean-shift algorithm takes a feature (range) bandwidth, spatial bandwidth, and a minimum region area (in pixels) as input. We need to adjust these three parameters for a good segmentation. In the graph-based algorithm [18], the predicate is based on both of the external and the internal intensity differences of the regions. More specifically, the external difference is calculated as the minimum weight edge between two neighboring regions and the internal difference of a region is the largest weight in the minimum spanning tree of that region. The predicate is then defined by comparing these two values to make an adaptive decision with respect to the local characteristics of an image. The graph-based algorithm [18] also contains three parameters, which are used to smooth the image, to control the minimum region size and to set the expected number of regions. All of these parameters are in different ranges, making the tuning of parameters tricky. As stated in Section III, the proposed DRM algorithm contains only one set of two parameters ($\lambda_1, \lambda_2$) in the consistency test. In implementation, we fix $\lambda_2$ to be 1, and hence only need to tune the value of $\lambda_1$. This makes DRM is much more convenient to control than the mean-shift algorithm [32] and the graph-based algorithm [18].

Figs. 10(b)-(d) show the segmentation results by the graph-based method [18], the mean-shift method [32] and the proposed DRM method. Segmentations are obtained by choosing the results with roughly the same number of regions from the three methods. By this setting, we can compare the performance of them in the same segmentation granularity. In Figs. 10(b)-(c), some major boundaries in the ground truth are missed, and some boundaries are over-detected. In Fig. 10(d), we can see that the proposed method can retain good localization of boundaries. In Fig. 10(e), the human labeled segmentation is used as the reference (ground truth) for visual evaluation of the segmentation quality. For each image, 5-10 segmentations produced by different human observers are chosen, and the consistency among different human segmentations is indicted by the probability-of-boundary ($P_b$) images (Fig. 10(e)), where the darker pixels means more people marked them as a boundary. For more information about the generation of human labeled ground truth segmentation, please refer to [37].

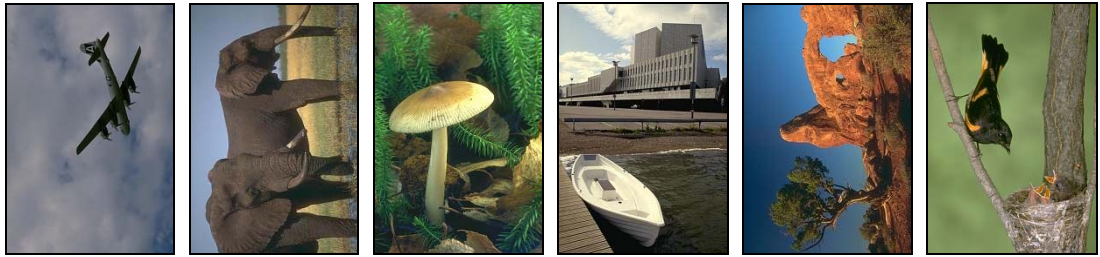

(a) Original images.

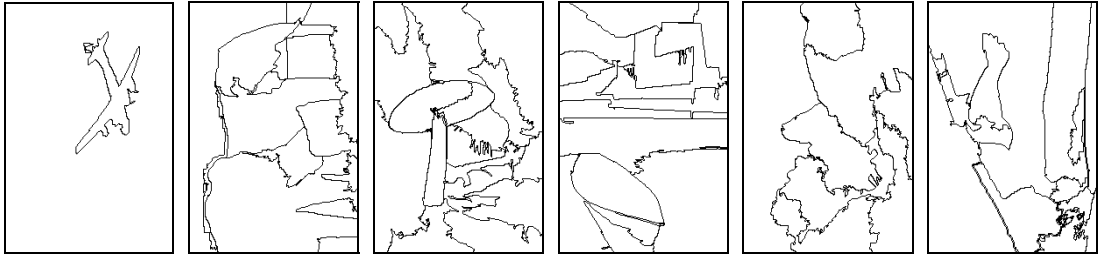

(b) Segmentation results by the graph-based region merging method in [18].

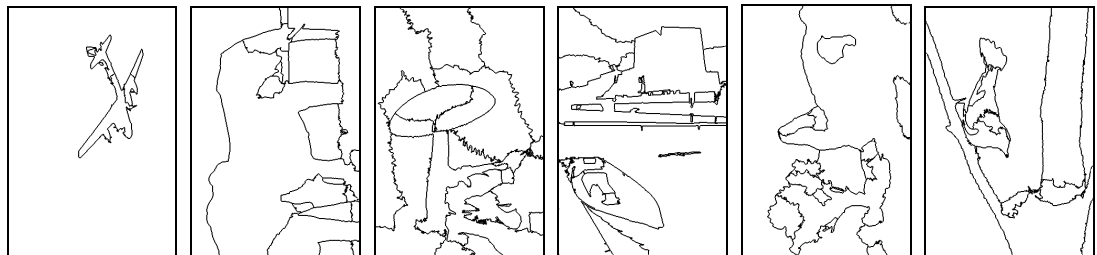

(c) Segmentation results of mean-shift method in [32].

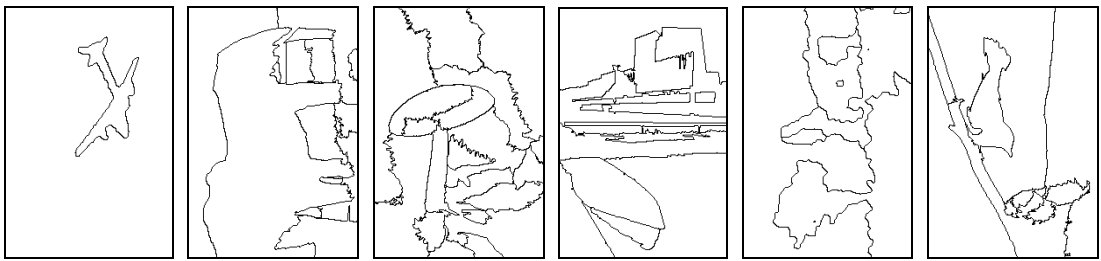

(d) Segmentation results of the proposed method with watershed initializations.

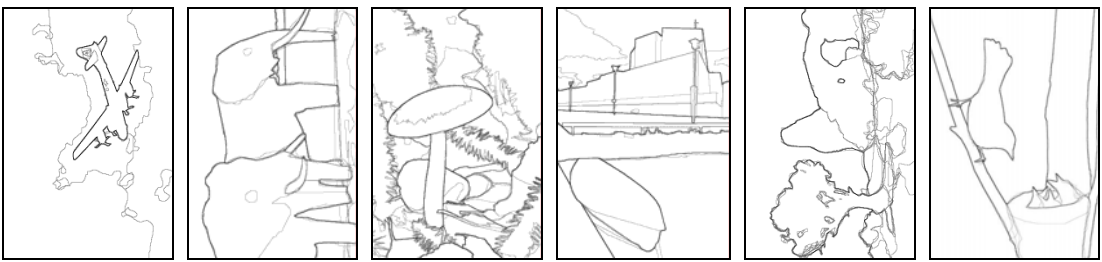

(e) Ground truth.

**Figure 10.** Segmentation results by different methods. The first row shows the original images. The second row shows the results by the graph-based image segmentation method [18]. The third row shows the results by the mean-shift method [32]. The forth row shows the proposed method. The fifth row shows probability-of-boundary images of the human segmentation, where the darker boundaries indicate more subjects marked a boundary [38].

Let's then discuss about the quantitative measure of segmentation quality. This task might be as hard as segmentation itself. In the past decades, many researchers have been studying the supervised evaluation methods, where the segmented images are compared with a ground truth image. However, in unsupervised image segmentation, often there is no ground truth available. In order to build a common platform for researchers to evaluate various image segmentation methods, a group of human segmented images for each test sample are provided in the Berkeley Segmentation Dataset (BSDS), and a boundary-based evaluation method was proposed by Martin *et al.* [38] on this dataset. In [38], the precision-recall framework is applied in conjunction with the human-marked boundaries. Precision-recall is a well-accepted measure technique in image segmentation, which considers two aspects of boundary qualities: *precision* is the fraction of detections that are true positives rather than false positives, while *recall* is the fraction of true positives that are detected rather than missed. A combination of these two quantities can be summarized as the *F*-measure in [39]:

$$F = PR/(\alpha R + (1-\alpha)P)$$

where $\alpha$ is a relative cost between *P* and *R*. An *F*-measure curve can be obtained by changing the algorithm parameters. Since the *F*-measure curve is usually unimodal, the maximal one may be taken as a summary of the algorithm's performance in the sense that large value of *F*-measure indicates high quality of image segmentation. In our experiment, the optimal parameter is chosen for each image and the corresponding *F*-measure is recorded for the overall evaluation.

There are two issues needed to address when use BSDS for evaluating the segmentation results. First, the ground truth is defined by a collection of 5 to 10 human segmentations for each image. Simply uniting the humans' boundary maps is not a reasonable choice because it ignores the high overlapping boundaries among different humans. Second, when matching boundary pixels, one should avoid over-penalizing the slightly mislocalized boundaries. We use the method proposed by Martin *et al.* to compute the *F*-measure for all algorithms. In this method, the segmentation results are compared with each human segmentation separately. False positives are counted as the boundary pixels that do not match any human boundary. The recall depends on all the human segmentations, i.e. averaged over different human data. A merit of Martin *et al.*'s method is that it can tolerate some localization errors in the matching process, which is a reasonable consideration in the correspondence of ground-truth and segmentation results. In Table 4, we compare the *F*-measures of the Canny edge detector [5], the graph-based method [18], the mean-shift method [32] and the DRM algorithm. The best parameter is chosen for each of these methods, and we have the *F*-measures as 0.62 for Canny edge detector, 0.62 for the graph-based method [18], 0.65 for mean-shift method [32] and 0.65 (with watershed as initial segmentation) and 0.66 (with mean-shift as initial segmentation) for the proposed DRM method.

**Table 4**. The best *F*-measures by the competing methods in the BSDS dataset. The average F-measure for the human subjects is 0.79 [40], which represents the human performance for the segmentation task. The DRM algorithm is tested when initialized by watershed and mean-shift respectively.

| Method | Human | Graph –based method [18] | Canny [25] | Mean-shift [32] | DRM (initialized by watershed) | DRM (initialized by mean-shift) |
|---|---|---|---|---|---|---|
| *F*-measure | 0.79 | 0.62 | 0.62 | 0.65 | 0.65 | 0.66 |

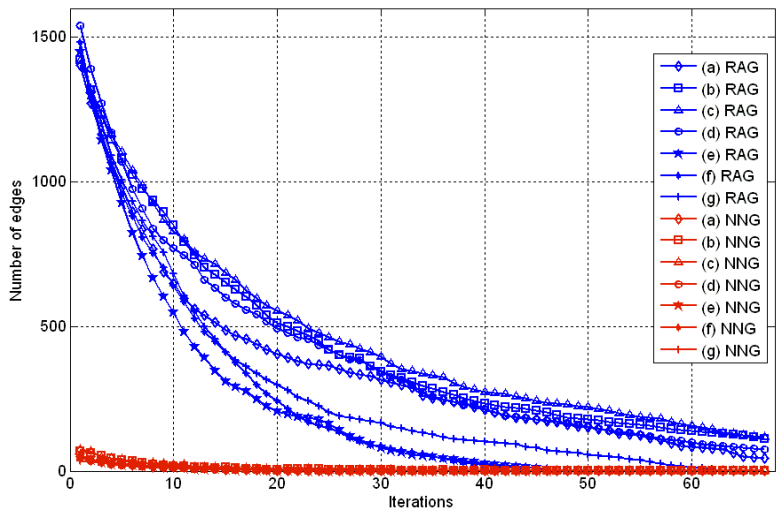

**Figure 11.** The number of RAG edges (in blue) vs. the number of NNG cycles (in red) for images in Figs. 9(a)-(g) at different region merging stages.

**VI-C. The performance of accelerated DRM algorithm**

In Fig. 11, we show the number of RAG regions and the number of NNG cycles for images Fig. 9(a)-(g) at different region merging stages. Apparently, the number of cycles (in red) is much smaller than that of the RAG edges (in blue). By using the accelerated region merging algorithm, the computational effort is largely saved. The update of NNG cycles depends on the size of RAG in the second order neighborhood. In Fig. 12, we show the histogram of RAG node degrees for the over-segmented images in Figs. 9(a)-(e) (in the second column). We can see that most of the nodes have the degree less than 15, therefore the computational cost for updating the NNG cycles is not expensive. This acceleration algorithm would be very helpful in the high-resolution environment where a massive graph containing billions of vertices will exist.

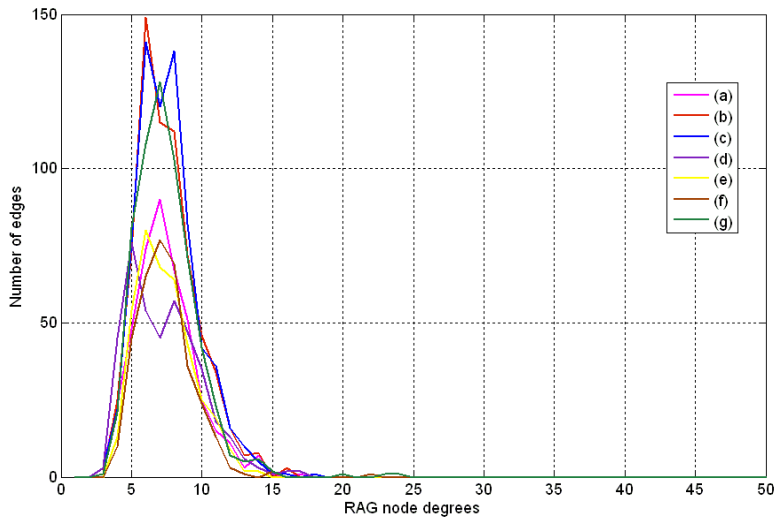

**Figure 12.** Histograms of the RAG node degrees for images in Figs. 9(a)-(g) (shown in different colors).

## VI-D. Discussions

The proposed DRM method checks the consistency between neighboring regions for region merging, and the region merging predicate uses the minimum weight edge between two regions in measuring the difference between them. It guarantees that some global properties are kept under this predicate. However, there can be a limitation for capturing the perceptual differences between two neighboring local regions. Another issue lies in the *DP* based merging order. Since decisions are made locally, DRM might be trapped into the local minimum, though it does possess some global properties as we proved in Section IV-B. Some failure examples are given in Fig. 13. We can see that the proposed DRM algorithm may miss some long but weak boundaries (1$^{st}$ example), and it may also merge the regions with short but high contrast boundary (2$^{nd}$ example).

There are several potential extensions to DRM to solve the above problems. For example, we can add a global refinement step to correct the miss-classified regions due to local decisions. Many merging errors in DRM are due to the insufficient local (perceptual) information used to making the merging decision. However, if we view the outputted labels of DRM as initial markers of the image content, then these initial markers can be used to calculate some global statistics of the image so that a global refinement can be defined to correct the DRM errors according to some criterion. Such a DRM with refinement scheme can exploit both local and global image features, and hence better results can be expected.

Another potential extension is the introduction of user interaction. With some user guidance, the initial label of part of the image regions can be assigned beforehand, which will provide useful information for the region merging process. An interactive DRM algorithm can then be developed to accomplish the image segmentation.

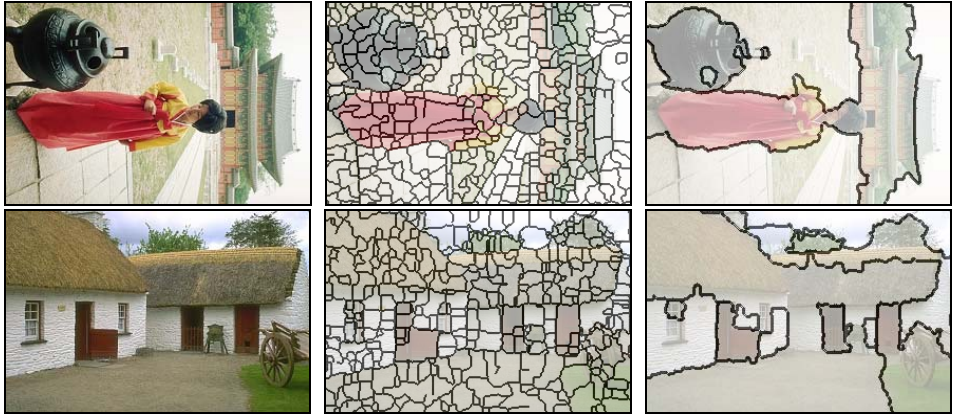

**Figure 13.** Failure examples of the proposed DRM method. The first column shows the original images, the second column shows the initial segmentation by watershed algorithm, and the third column shows the region merging results by the proposed method. We see that DRM methods may miss some weak boundaries (in the 1st example) and merge the regions with short but high contrast boundary (in the 2nd example).

## VII. Conclusions

In this paper, we proposed a novel method for segmenting an image into distinct components. The proposed algorithm is implemented in a region merging style. We defined a merging predicate $P$ for the evidence of a merging between two neighboring regions. This predicate was defined by the sequential probability ratio test (SPRT) and the maximum likelihood criterion. A dynamic region merging (DRM) was then presented to automatically group the initially over-segmented many small regions. Although the merged regions are chosen locally in each merge stage, some global properties are kept in the final segmentations. For the computational efficiency, we introduce an accelerated algorithm by using the data structure of region adjacency graph (RAG) and nearest neighbor graph (NNG). Experiments on natural images show the efficiency of the proposed algorithm. There are several potential extensions to this work, such as the introduction of global refinement and user interaction, etc. Those will be further investigated in our future work.

# References

[1] D.A. Forsyth and J. Ponce, Computer Vision: A Modern Approach. Prentice Hall, 2002

[2] L. Ladicky, C. Russell, P. Kohli, P. Torr. Associative Hierarchical CRFs for Object Class Image Segmentation. In: ICCV 2009.

[3] F, Lecumberry, A, Pardo and G. Sapiro. Simultaneous object classification and segmentation with high-order multiple shape models. IEEE Transactions on Image Processing. pp: 625 - 635, 2010.

[4] R.C. Gonzalez and R.E. Woods. Digital Image Processing. Addison Wesley, Reading, MA, 1992.

[5] J. Canny. A Computational Approach to Edge Detection, IEEE Trans. Pattern Analysis and Machine Intelligence, vol. 8, pp. 679-698, 1986.

[6] B. Paul, L. Zhang and X. Wu, "Canny edge detection enhancement by scale multiplication," *IEEE. Trans. on Pattern Analysis and Machine Intelligence*, vol. 27, pp. 1485-1490, Sept. 2005.

[7] L. Zhang, B. Paul, et al, "Edge detection by scale multiplication in wavelet domain," *Pattern Recognition Letters,* vol. 23, pp. 1771-1784, 2002.

[8] J. Shi and J. Malik. Normalized Cuts and Image Segmentation. IEEE Transactions on Pattern Analysis and Machine Intelligence (PAMI) 2000.

[9] S. Wang, J. M. Siskind. Image Segmentation with Ratio Cut, IEEE Transactions on Pattern Analysis and Machine Intelligence, 25(6):675-690, 2003.

[10] Z. Wu and R. Leahy. An optimal graph theoretic approach to data clustering Theory and its application to image segmentation. IEEE Transactions on Pattern Analysis and Machine Intelligence. November 1993.

[11] H. D Cheng, Y. Sun. A hierarchical approach to color image segmentation using homogeneity. IEEE Transactions on Image Processing. Volume: 9 , Issue: 12, page(s): 2071-2082, 2000.

[12] S. Lee; M.M. Crawford. Unsupervised multistage image classification using hierarchical clustering with a bayesian similarity measure. IEEE Transactions on Image Processing. Page(s): 312 -320, 2005.

[13] A. Moore, S. J. D. Prince, J. Warrell, U. Mohammed, and G. Jones. Superpixel lattices. CVPR, 2008

[14] A. Moore, S. J. D. Prince, J. Warrell, U. Mohammed, and G. Jones. Scene shape priors for superpxiel segmentation. ICCV, 2009.

[15] A. Moore, S. Prince . "Lattice Cut" - Constructing superpixels using layer constraints. CVPR 2010.

[16] R. Nock and F. Nielsen. Statistic region merging. IEEE Trans. on Pattern Analysis and Machine Intelligence, vol 26, pages 1452-1458, 2004.

[17] K. Haris and S. N. Estradiadis and N. Maglaveras and A. K. Katsaggelos. Hybrid image segmentation using watersheds and fast region merging. IEEE Transactions on Image Processing, vol 7, pp. 1684-1699, Dec. 1998.

[18] P.F. Felzenszwalb and D.P. Huttenlocher. Efficient Graph-Based Image SegmentationInternational Journal of Computer Vision. Vol. 59, Number 2, September 2004.

[19] B. Peng, L. Zhang and J. Yang, Iterated Graph Cuts for Image Segmentation. In Asian Conference on Computer Vision, 2009.

[20] Moscheni, F. Bhattacharjee, S. Kunt, M. Spatio-temporal segmentation based on region merging. IEEE Transactions on Pattern Analysis and Machine Intelligence. Vol. 20, pages: 897-915. Sep 1998.

[21] J. Ning, L. Zhang, D. Zhang and C. Wu. Interactive Image Segmentation by Maximal Similarity based Region Merging. Pattern Recognition, vol. 43, pp. 445-456, Feb, 2010

[22] F. Calderero, F. Marques. Region merging techniques using information theory statistical measures. IEEE Transactions on Image Processing. Volume: 19 , Issue: 6. page(s): 1567-1586, 2010.

[23] F. Calderero, F. Marques. General region merging approaches based on information theory statistical measures. The 15th IEEE International Conference on Image Processing (ICIP). pp: 3016-3019, 2008.

[24] H. Liu, Q. Guo, M, Xu, I. Shen. Fast image segmentation using region merging with a k-Nearest Neighbor graph. IEEE Conference on Cybernetics and Intelligent Systems. Page(s): 179-184, 2008.

[25] Y. Shu, G. A. Bilodeau, F. Cheriet. Segmentation of laparoscopic images: integrating graph-based segmentation and multistage region merging. The 2nd Canadian Conference on Computer and Robot Vision, 2005.

[26] I.E. Gordon, Theories of Visual Perception, first ed. John Wiley and Sons, 1989.

[27] K. Koffka. Principles of Gestalt Psychology. New York: Harcourt, Brace and World. 1935.

[28] A. Wald. Sequential Analysis. Wiley Publications in Statistics, Third ed. Wiley, 1947.

[29] Z. Wu, Homogeneity testing for unlabeled data: A performance evaluation, CVGIP: Graph. Models Image Process., vol. 55, pp. 370-380, Sept. 1993.

[30] A. Trémeau and P. Colantoni. Regions adjacency graph applied to color image segmentation. IEEE Transactions on Image Processing. Vol. 9, pp. 735-744, 2000.

[31] L. Vincent, P. Soille, Watersheds in digital spaces: an efficient algorithm based on immersion simulations, IEEE Transactions on Pattern Analysis and Machine Intelligence 13 (6) 583–598, 1991.

[32] D. Comanicu, P. Meer. Mean shift: A robust approach toward feature space analysis. IEEE Trans. on Pattern Analysis and Machine Intelligence, 24, 603-619, May 2002.

[33] Bellman, Richard, Dynamic Programming, Princeton University Press, 1957.

[34] A. Mishra, P. Fieguth, D. Clausi. Accurate boundary Localization using Dynamic Programming on Snakes. Proceedings of the Canadian Conference on Computer and Robot Vision, pp 261-268, 2008.

[35] A. A. Amini, T.E. Weymouth, and R.C. Jam. Using dynamic programming for solving variational problems in vision. IEEE Trans. on Pattern Analysis and Machine Intelligence, 12(9):855-867, 1990.

[36] D. Eppstein, M. S. Paterson, and F. F. Yao. On nearest-neighbor graphs. Discrete & Computational Geometry 17(3):263–282, Apr 1997.

[37] D. Martin, C. Fowlkes, D. Tal, and J. Malik. A Database of Human Segmented Natural Images and its Application to Evaluating Segmentation Algorithms and Measuring Ecological Statistics. ICCV 2001.

[38] D. Martin, C. Fowlkes, and J. Malik. Learning to detect natural image boundaries using local brightness, color and texture cues. IEEE Trans. on Pattern Analysis and Machine Intelligence, 26(5):530–549, 2004.

[39] C. Van Rijsbergen. Information Retrieval, second ed. Dept. of Computer Science, Univ. of Glasgow, 1979.

[40] P. Arbelaez, M. Maire, C. Fowlkes, and J. Malik. From Contours to Regions: An Empirical Evaluation. IEEE Conference on Computer Vision and Pattern Recognition, 2009.